\documentclass[sigconf, 9pt, nonacm]{acmart}
\AtBeginDocument{%
  }

\setcopyright{acmlicensed}
\copyrightyear{2024}
\acmYear{2024}
\acmDOI{XXXXXXX.XXXXXXX}

\acmConference[Conference acronym 'XX]{Make sure to enter the correct
  conference title from your rights confirmation emai}{June 03--05,
  2018}{Woodstock, NY}
\acmISBN{978-1-4503-XXXX-X/18/06}

\usepackage{blindtext}
\usepackage{siunitx}
\usepackage{amsmath}
\usepackage{nicefrac}
\usepackage{subfig}

\newcolumntype{M}[1]{>{\centering\arraybackslash}m{#1}}
\newcolumntype{R}[1]{>{\raggedleft\arraybackslash}m{#1}}
\newcolumntype{L}[1]{>{\raggedright\arraybackslash}m{#1}}
\graphicspath{ {./images/} }

\begin{document}

\title{FLAME: Adaptive and Reactive Concept Drift Mitigation for Federated Learning Deployments}

\author{Ioannis Mavromatis}
\orcid{1234-5678-9012}
\affiliation{%
  \institution{Digital Catapult}
  \city{London}
  \country{UK}
}
\email{ioannis.mavromatis@digicatapult.org.uk}

\author{Stefano De Feo}
\affiliation{%
  \institution{Dept. of Electrical and Electronic Engineering, University of Bristol}
  \city{Bristol}
  \country{UK}}

\author{Aftab Khan}
\affiliation{%
  \institution{Bristol Research \& Innovation Laboratory,  Toshiba Europe Ltd.}
  \city{Bristol}
  \country{UK}
}
\email{aftab.khan@toshiba-bril.com}

\renewcommand{\shortauthors}{I. Mavromatis et al.}

\begin{abstract}
This paper presents Federated Learning with Adaptive Monitoring and Elimination (FLAME), a novel solution capable of detecting and mitigating concept drift in Federated Learning (FL) Internet of Things (IoT) environments. Concept drift poses significant challenges for FL models deployed in dynamic and real-world settings. FLAME leverages an FL architecture, considers a real-world FL pipeline, and proves capable of maintaining model performance and accuracy while addressing bandwidth and privacy constraints. Introducing various features and extensions on previous works, FLAME offers a robust solution to concept drift, significantly reducing computational load and communication overhead. Compared to well-known lightweight mitigation methods, FLAME demonstrates superior performance in maintaining high F1 scores and reducing resource utilisation in large-scale IoT deployments, making it a promising approach for real-world applications.
\end{abstract}




\maketitle


\section{Introduction}
The Internet of Things (IoT) is becoming a standard solution for enhancing and maintaining services and applications in both industrial and non-industrial environments. IoT devices play a critical role in healthcare~\cite{kodali2015implementation}, manufacturing~\cite{zhong2018internet}, product life cycles and warehouse inventory management~\cite{tejesh2018warehouse}, among others. 

All IoT systems must meet real-time performance requirements while adhering to constraints in power consumption, physical size, installation complexity and more~\cite{capra2019edge}. IoT devices face data privacy concerns~\cite{privacy_in_ml_challenges}, communication bandwidth limitations~\cite{10380759}, processing limitations, etc. Machine Learning (ML) is very prominent in IoT, used in multiple scenarios to forecast future events and behaviours, optimise various tasks and solves many of the problems and challenges mentioned above~\cite{iotMLSurvey}. Bandwidth limitations and privacy concerns are prominently addressed by distributed ML architectures, such as Federated Learning (FL)~\cite{privacy_in_ml_challenges}.

The dynamic nature of a real-world IoT environment can lead to drastic changes in the input data distributions~\cite{le3dDataDrift}. From the ML model's perspective, the underlying relationship between the input data and expected output (target) changes over time, leading to an underfitted model. This behaviour is called concept drift~\cite{smartcities4010021}. It can occur for several reasons, e.g., long-term changes in the data (e.g., environmental changes), faulty hardware (sensor drift), or even adversarial actions, such as data poisoning attacks. These scenarios can be detrimental to ML predictions' quality when considering large-scale IoT systems with misbehaving ``production'' models.

This is the problem we consider in this paper. We present Federated Learning with Adaptive Monitoring and Elimination (FLAME) of concept drift. FLAME provides an automated and adaptable way of detecting and handling concept drift within an FL-based learning and inference pipeline. Within such a pipeline and post-deployment, it is important to identify when a model requires retraining without saturating the system resources and the data to be used for high accuracy of the ML predictions over time.

FLAME operates within an FL setup. It captures the entire lifecycle of an ML model, considering the stability of the initial training, its continuous performance and the mitigation strategies against drift, all these implemented in a real-world three-tier architecture, i.e., ``cloud-edges-microcontrollers''. Due to the nature of the embedded microcontrollers (i.e., minimal processing power), FLAME considers that edge devices installed close to the IoT sensors are used for the training, the cloud is used for the aggregation of a global model, and the model is later converted to its embedded form and deployed to the microcontrollers for inference~\cite{flare}.

As bandwidth and computing limitations are paramount for IoT deployments, we evaluate FLAME not only on the perceived accuracy of the models over time but also on the data exchange introduced and the need for resource-intensive retraining, ensuring it is always kept to the minimum. Moreover, we consider an unlabelled data scenario, representing a realistic case for an IoT system. We compare FLAME against various traditional statistical approaches, e.g., Adaptive Windowing (ADWIN)~\cite{adwin}, Kolmogorov-Smirnov Windowing (KSWIN)~\cite{kswin}, etc., frequently used from IoT deployments due to their lightweight nature.

The remaining of the paper is structured as follows. Sec.~\ref{sec:related_work} presents similar works identified in the literature. A typical system architecture and how FLAME can operate within an FL pipeline is described in Sec.~\ref{sec:architecture}. The enhancements introduced within our solution, the model and dataset used and the algorithms compared against FLAME are outlined in Sec.~\ref{sec:flame}. Finally, our results are presented in Sec.~\ref{sec:results} and our final remarks are summarised in Sec.~\ref{sec:conclusions}.


\section{Related Work}\label{sec:related_work}
Based on the data distribution changes, concept drift is classified as abrupt, incremental, gradual or recurring~\cite{le3dDataDrift}. Many works present centralised mitigation strategies for resource-constrained environments. For example, a lightweight concept drift detector was presented in~\cite{lightWeightConceptDrift}. The authors calculate the centroids of each data class for old and new data, and finding the distance between the centroids can detect whether drift occurred. For unsupervised detection, such as in our case, the authors rely on a $k$-means clustering approach to label the data. This approach may work well for simplistic datasets but will underperform in more complex scenarios like the one introduced in this paper. In~\cite{iotDriftDetection}, authors detect drift using Principal Component Analysis and, after removing the outliers, develop a dynamic way of changing the depth of the ML model to adapt to the new data distribution. Even though this approach may lead to great performance over time and tackle problems such as forgetfulness, the model's variable size will make the model deployment in resource-constrained devices rather prohibitive.

Research around distributed IoT environments is very limited. In~\cite{smartcities4010021}, four well-known drift detection methods were successfully implemented within a distributed architecture, proving that detection is not negatively impacted by distributing a system over a network. Concept drift detection and mitigation become harder in distributed environments due to the high communication and computing costs required for retraining. In distributed architectures where many models are trained separately, drift adaptation can be enabled by models from other systems where the drift was detected (leveraging the asynchronous appearance of concept drifts). For example, in~\cite{HAng2013}, the classifiers on a given node consist of an ensemble of classifiers trained on other nodes after drift is detected. Similar approaches are seen in federated systems such as~\cite{GCanonaco2021,FCasado2021}, capitalising on models from other clients through model aggregation. However, both works become rather impractical for a real-world distributed scenario due to the increased volume of resources they consume. The solution in~\cite{GCanonaco2021} uses continuous learning, constantly exchanging the model parameters and data, and~\cite{FCasado2021}, even though it reacts to concept drift better with reduced communications and memory usage, it stores all raw data from previous concepts significantly increasing the storage requirements on the edge devices.

\begin{figure*}[t]
    \centering
    \includegraphics[width=0.83\textwidth]{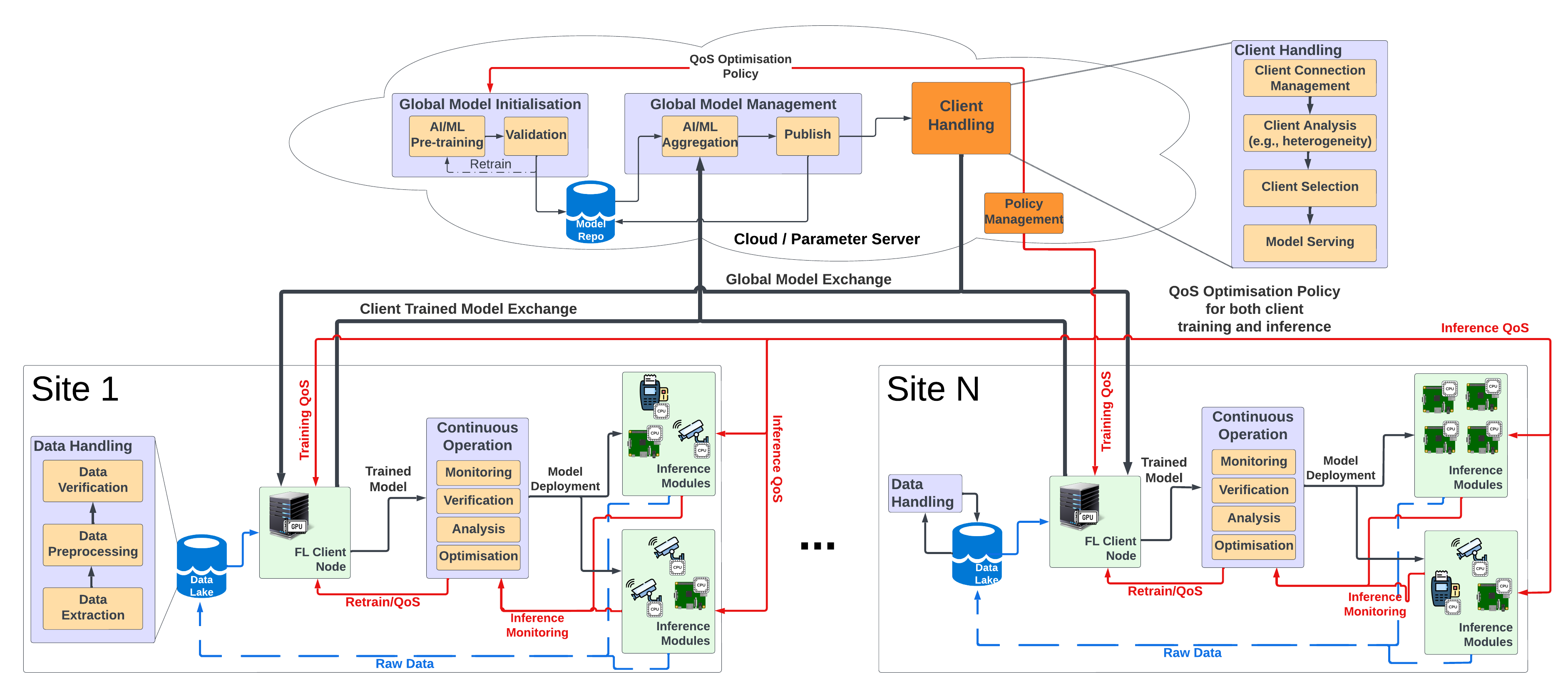}
    \caption{Overview of a typical FL lifecycle in FedOps for our scenario. It is distinguished in the ``cloud'' (handling the parameter server) and the ``on-site'' (handling the client and sensor nodes) deployments.}
    \label{fig:FedOps}
\end{figure*}


\section{System Architecture}\label{sec:architecture}
We consider a practical application of a smart city scenario. The system comprises various devices such as surveillance cameras, traffic sensors, smart meters, public Wi-Fi hotspots, etc. As discussed, our system operates in a three-tier architecture, i.e., cloud-edge-endpoints, where each tier provides different computing capabilities. All IoT devices (i.e., namely, the endpoints) frequently receive firmware updates that solve bugs or add new functionality. Moreover, these IoT devices can continuously monitor their system logs, network traffic patterns and application behaviours for security breaches, malware, or abnormalities in their operation. 

An FL deployment in such a system can utilise the more powerful edge devices to train and validate ML models. The cloud can aggregate a global model, and the trained models are deployed in the existing IoT devices for inference. Such a setup enables data privacy and security as the inference data remain on the local IoT devices, ensures the system's scalability and adaptability, allowing the system to update its detection models continuously, and also achieves reduced latency and bandwidth usage as the data do not need to be exchanged over low-datarate links.

\vspace{-3mm}

\subsection{FedOps for Large Scale IoT Deployments}
The above FL deployment needs to consider the entire lifecycle of the ML pipeline, usually divided into 4 phases -- scoping, data, modelling, and deployment. Machine Learning Operations (MLOps) refers to the practices, tools and techniques that can realise and manage the ML model lifecycle from the design and training to the evaluation, distribution and deployment. Federated Learning Operations (FedOps) extends the concept of MLOps within the FL space. FedOps considers the distributed nature of an FL deployment and orchestrates the data exchange between the clients, how and when models are aggregated, how models are initialised and monitored and how they can be extended with personalisation to fit heterogeneous deployments of diverse devices~\cite{moon2024fedops}.

The example FedOps deployment in~\cite{moon2024fedops} selects FL clients based on their communication cost and overall model accuracy. This work does not consider model monitoring, making their implementation impractical for a real-world IoT system. We build upon that and introduce FLAME within a realistic FedOps pipeline (Fig.~\ref{fig:FedOps}). During training, the model weights are initiated from the parameter server. The local datasets across all clients are processed and prepared for training. Each client splits the dataset into three train-test-validate splits (e.g., we use $80\%-10\%-10\%$, respectively). In Sec.~\ref{subsec:dataset}, we describe our dataset preprocessing and the features extracted from that. A FedOps pipeline will later incorporate a hyperparameter tuning phase. For our experiments, during this phase, we fine-tuned the ML and statistical models hyperparameters, as described in Secs.~\ref{subsec:detectors} and~\ref{subsec:mlmodel}. Following, during training, an ML model is usually trained until a stability point is reached. For our experimentation, we randomly initialised the set of hyperparameters and trained the model until the stability point, introducing a hard-coded epoch threshold to disregard underperforming attempts. Sec.~\ref{sec:results} presents the results of a successful pipeline run. The trained model is later deployed for inference to the endpoint nodes. Each endpoint uses its locally collected data as an input for the inference phase.

During training and inference, the model's performance is continuously monitored for concept drift according to policies introduced by the parameter server. When a model is tagged as ``no-longer-optimal,'' it can be retrained or retired and replaced by a new one. For both mitigation approaches, endpoint data should be sent to the client nodes, verified for validity, and used for training. Our solution enables the above functionality with two schedulers and a ``concept''-aware dataset creation as described in Sec.~\ref{subsec:enhancements}. Finally, a production FedOps system will usually incorporate various event handlers and result/log collection mechanisms responsible for monitoring requirements and triggering various system events.


\section{FLAME: Concept Drift Detection Pipeline}\label{sec:flame}
We implement our solution within the above pipeline, extending our previous work Federated LeArning with REactive monitoring of concept drift (FLARE)~\cite{flare}. FLARE incorporated two scheduling subsystems, one for the FL clients and one for the endpoints. The client's scheduler is responsible for monitoring the model training and assessing the model's stability, i.e., when it is ready for inference. A stable model is converted into its embedded form and is sent to the endpoint nodes. The stability point is found by measuring the difference between the validation loss (using a validation dataset) and the training loss (using the training dataset) across different time windows. A model is considered stable when the $\sigma_w < \sigma_s \times (1 - \beta)$  holds. In this equation, $\sigma_w$ is the standard deviation in the current window, and $\sigma_s$ is the previous stable standard deviation value. $\beta$ represents a stability coefficient, where a higher $\beta$ increases the sensitivity to concept drift and the communication cost. 

The endpoint scheduler runs a confidence validation test using the Kolmogorov-Smirnov (KS) test~\cite{ksTest}, calculating values between $0$-$1$ and comparing the confidence from the client's validation set against the current dataset on the endpoint. When the similarity is low (KS-test result close to $1$), this indicates a change in the sensor data distribution and the latest data are sent to the client for further training. On the other hand, a high similarity (KS-test result close to $0$) indicates the deployed model is still effective on the current dataset. A static threshold $\phi$ was used to determine whether the similarity was high or low. More information can be found in~\cite{flare}.

\begin{figure}[t]
    \centering
    \includegraphics[width=0.98\columnwidth]{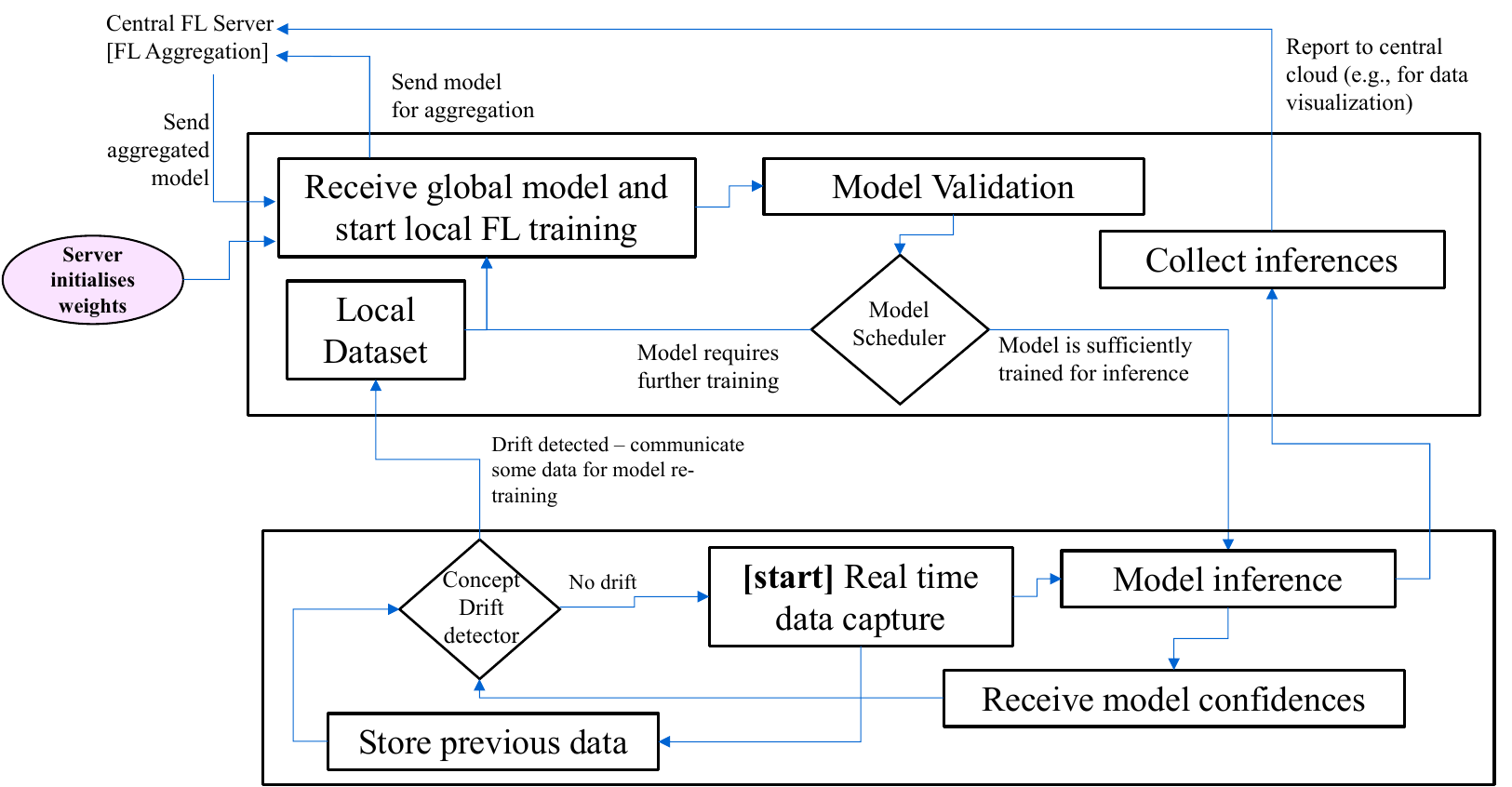}
    \caption{A system diagram showing the operation of FLAME.}
    \label{fig:system_diagram}
\end{figure}


\subsection{FLAME Enhancements}\label{subsec:enhancements}

FLAME considered the intricacies of a real-world IoT system, such as the limited communication bandwidth, the resource-sparse nature of the devices, etc., and the long-term continuous operation within an IoT platform. Fig.~\ref{fig:system_diagram} shows the interactions across the various system components. FLAME extends FLARE's functionality in three ways. First, we introduce a dynamic threshold $\phi_\mathrm{a}$ to compare the KS test. The adaptive thresholding technique takes all KS statistics' mean $\mu_\mathrm{a}$ and standard deviation $\sigma_\mathrm{a}$ within a variable-sized window $w_\mathrm{a}$ and sets $\phi_\mathrm{a} = 3\sigma_\mathrm{a} + \mu_\mathrm{a}$. The window $w_\mathrm{a}$ is defined as the last $n$ KS test values.

The variable-sized window is designed to ensure the threshold is based only on relevant previous KS statistics, allowing for the detection of different types of drift. To achieve this, only the KS statistics recorded since the last drift detection are considered, as earlier concepts should not influence the current threshold. Additionally, even within the same concept, older KS statistics can distort the threshold and should eventually be marked as stale and excluded. In the proposed method, the oldest \nicefrac{1}{3} of the confidence values are discarded from the window before the KS test calculation. This windowing strategy is effective for long-term deployments as the increasing window size enhances the threshold's reliability.

\begin{figure}[t]
    \centering
    \subfloat[\centering Model stability.\label{fig:stability}]{{\includegraphics[width=0.4\columnwidth]{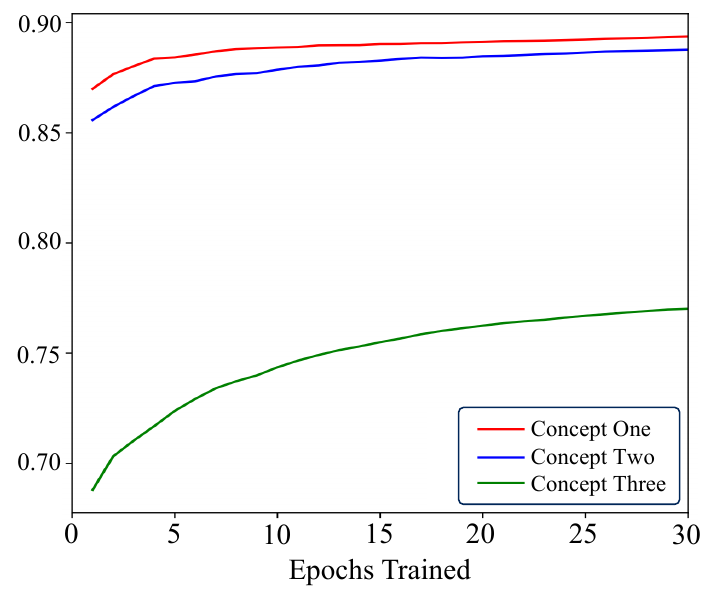} }}
    \qquad
    \subfloat[\centering Concept Examples.\label{fig:concepts}]{{\includegraphics[width=0.5\columnwidth]{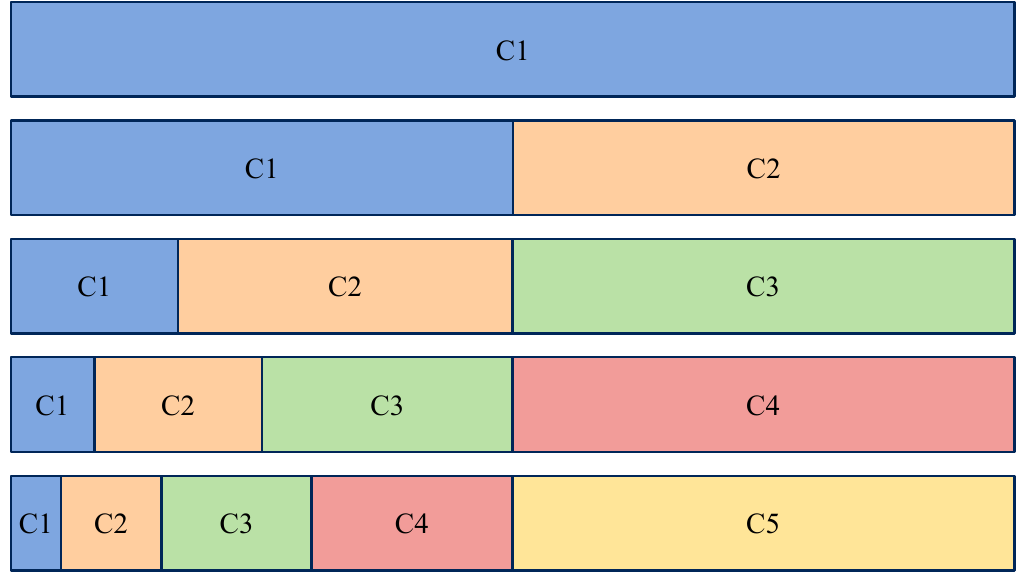} }}
    \caption{a) Stability example for a model training on three different concepts in the MNIST-C dataset~\cite{mnistC}, b) an example of five concepts and the split of the newly created dataset.}
    \label{fig:mnist}
\end{figure}

Moreover, the static threshold $\beta$ used for the model's stability is enhanced with an adaptive solution. The stability measurements across different concepts may vary significantly, with some concepts never reaching the threshold $\beta$, thus making the method very sensitive (Fig.~\ref{fig:stability}). Instead of $\beta$, we use the gradient of the stability curve for a given concept (as opposed to an absolute value), leading to a robust and adaptive method. When the gradient increases above a specific value, the model can be marked as stable. This allows us to work with imbalanced classes and datasets where one class may underperform compared to the rest.

As a final improvement, we introduce the retention of training data from previous concepts (Fig~\ref{fig:concepts}). FLARE, when $N$ new samples were received for training, the $N$ oldest samples were removed from the training dataset before training, resulting in a dataset with a constant size. However, this can be catastrophic over time, with the model forgetting older concepts. Our new approach picks random samples from all concepts based on the following. Let $\mathcal{C} \triangleq \left \{C_1,\ldots , C_n \right \}$ with $n \in \mathbb{N}^*$ denoting all concepts that have appeared in the system where $C_n$ is the newest concept. Also, let $\mathcal{S}_{C_n}$ be all the samples of the concept $C_n$. Then, the total dataset $D$ consists of:
\begin{align}
D &= D_n \cup D_{n-1} \cup \ldots \cup D_1 , \\
D_n & \subseteq \mathcal{S}_{C_n}, \quad \mathrm{where} \, |D_n| = \left\lfloor \frac{1}{2} |\mathcal{S}_{C_n}| \right\rfloor , \\
D_{x} &\subseteq \mathcal{S}_{C_x} \quad \mathrm{where} \, |D_{x}| = \left\lfloor \frac{1}{2} \frac{x \, |\mathcal{S}_{C_x}|}{\sum_{i=1}^{n-1} i} \right\rfloor, \, \forall x \in [1, n-1]
\end{align}
resulting in $50\%$ of the samples coming from the newest concept and smaller subsets from the older ones. Following such an approach, we can ensure that the model will perform best on the newest data without forgetting the older ones. Moreover, the training, validation, and test datasets extracted are always kept constant in size without draining the resources of the edge and sensor nodes.

\subsection{Concept Drift Detection Algorithms}\label{subsec:detectors}
Many lightweight concept drift detectors are found in the literature. We chose three detectors to compare our method with, these being ADWIN~\cite{adwin}, PHT~\cite{pageHinkley}, and KSWIN~\cite{kswin}. These detectors were chosen as they have been frequently used in the literature for similar concept drift detection activities~\cite{smartcities4010021} and operate comparably well on unlabelled data distributions~\cite{le3dDataDrift}.

\subsubsection{ADaptive WINdowing (ADWIN) algorithm}

ADWIN can detect distribution changes and drifts in data that vary with time. It uses an adaptive sliding window recalculated online according to the rate of change observed from the data. The window is discretised in two sub-windows without overlap. When a new sample is received, ADWIN examines all possible window splits, calculating the mean values for both windows and their absolute difference. The optimal lengths for the two sub-windows are found based on a threshold compared against all the calculated values. 

Once a drift is detected, all the old data samples within the first window are discarded. ADWIN can effectively detect gradual drift since the sliding window can be extended to a large-sized window and identify long-term changes. Abrupt changes can again be identified with a small number of samples due to the big difference introduced in the mean values.

\subsubsection{Page-Hinkley Test (PHT) algorithm}

PHT is a variant of the CUmulative SUM (CUSUM) test. It has optimal properties in detecting changes in the mean value of a normal process. PHT's two-sided extension was considered in this work~\cite{le3dDataDrift}. PHT recalculates the mean value and cumulative sum for every sample received based on a user-defined value.

PHT compares the mean values against a change detection threshold. The user-defined value's magnitude controls PHT's tolerance, while the change detection threshold tunes the false alarm rate. Larger thresholds entail fewer detections of false positives while increasing the number of false negatives. PHT easily identifies abrupt drifts due to the sudden change in the mean value. In contrast, incremental drift can be identified by sporadically sampling the time-series data stream.

\subsubsection{Kolmogorov-Smirnov Windowing (KSWIN) algorithm}

KSWIN is based on a KS test that accepts one-dimensional data and operates without assuming the underlying data distribution. KSWIN maintains a fixed-size sliding window discretised in two sub-windows without overlap. A two-sampled KS test is performed on both sub-windows. It compares the absolute distance between two empirical cumulative data distributions.

The test result is compared against the square root of the log of a sensitivity parameter over the length of the window. Data with increased periodicity and a large window make KSWIN too sensitive and return many false positives. Relatively small window sizes and an optimised sensitivity parameter significantly improve the performance. As described in~\cite{kswin}, KSWIN can detect gradual and abrupt drifts but falsely classifies many samples as false positives. However, considering the criticality of an error, false positives are not as critical and can be removed in post-processing.



\section{Results}\label{sec:results}
Our experimental setup consists of a ``cloud'' node, which is used as the model aggregator, 8 ``edge'' nodes that act as the FL clients, and 32 ``endpoint'' nodes, the embedded microcontrollers used for inference and ``raw data collection''. The training dataset (Sec.~\ref{subsec:dataset}) was split equally across the different clients, randomly allocating $\sim42$k applications to each. Each endpoint was assigned $\sim70$k for inference, splitting the available applications equally. Our experiment runs chronologically (monthly), so a sensor infers or a client trains/validates, respectively, on the sub-dataset of the given month. Finally, out of the 92 months of the entire dataset's length, the first 12 months (year 2014) are used for the initial model training and the remaining 80 months (Jan. 2015 - Sep. 2021) for inference. 

For the fine-tuning of the different detectors, we initially ran a grid search to estimate the value for all hyperparameters. The detectors' sensitivity was later fixed for the rest of the experiment. We chose the values that did not return candidate drift within the ``training'' and ``validation'' sets used for the initial 12 months of training. Finally, for our ML model, we used the FedAvg aggregation method, cross-entropy loss, ADAM optimisation, a learning rate of 0.003 and a batch size of 4. Finally, some results will be described only in text due to the limited space.


\subsection{Malware Detection Dataset}\label{subsec:dataset}
Our experimentation is based on the Androzoo dataset~\cite{androzoo}.  Androzoo contains millions of malware scans for different Android apps since 2010. The dataset was chosen due to the real and virtual concept drifts presented -- the types of malware change over time (virtual drift) and the efficacy of virus scanners (real drift).

Each app is scanned by over 70 antivirus scanners using VirusTotal\footnote{VirusTotal: https://www.virustotal.com/}, and Androzoo reports the number of times it is classified as malware. This score is used as the ground truth label for each app. As discussed in~\cite{NZakeya2021}, no definitive threshold classifies an app as malware; therefore, for our experimentation, a threshold of two detections was chosen to avoid single-case outliers. This threshold has also been used in previous works~\cite{DArp2014}. We used around $2.6\mathrm{M}$ applications for our experiment, with the percentage of malware at around $10\%$. This is considered realistic in terms of the proportion of real-world apps that contain malware, according to \cite{FPendlebury2019}. 

The Androzoo dataset consists of the full APK file for every app, so all APK files were pre-processed to extract the features required. Two main types of raw features can be extracted from the apps, i.e., raw opcode sequences or bytecode~\cite{surveyAndroozoo}. Opcode sequences were chosen due to their successful prior use in~\cite{NMcLaughlin2017}. They were extracted using Apktool\footnote{Apktool: https://ibotpeaches.github.io/Apktool/}, similarly with~\cite{NMcLaughlin2017}. As this work presents, a high F1 Score can be achieved by classifying these opcode sequences, so this was considered a reasonable method for classification.


\subsection{ML Model}\label{subsec:mlmodel}
We chose a simple Convolutional Neural Network (CNN) architecture that can be easily converted and deployed on embedded devices while retaining high accuracy. The model consists of an embedding layer, one convolutional layer followed by a max pooling layer, one hidden connected layer (fully connected) and a softmax classification layer (fully connected). ReLU activation functions were used on the convolutional layer, and the fully connected layer and a sigmoid function were applied to the softmax layer to produce the positive class probability. 

The opcodes are projected in the embedding space, and each vector is multiplied by a weight matrix that is initially randomised and updated by back-propagation during training. By doing so, the semantic information of each opcode can be encoded in the embedded space, and the network can discover certain opcodes with similar meanings, allowing them to be treated comparably by deeper layers in the network. The kernel size chosen for the convolutional layer was 64, which means it can form features for sequences of up to 64 opcodes; this is a reasonable choice given that the opcode sequence length was 357. Moreover, before applying the convolutional filters, we zero-pad the start and end of the input to ensure that the length of the output matrix from the convolutional layer is the same as the length of its input. Finally, the max pooling layer takes a maximum over the sequence length dimension, which allows sequences of different lengths to be represented by feature vectors of the same length. For more details, refer to~\cite{NMcLaughlin2017}.

The class imbalance (roughly $90\%$ benign - $10\%$ malware), was tackled with a weighted loss function to ensure no bias towards the benign (majority) class. Binary cross entropy was used with an increased weight to the malware class, with the weights calculated as the inverse of the square root of the number of samples per class.


\subsection{Preliminary FL Experiment}
\begin{figure}[t]
    \centering
    \includegraphics[width=0.98\columnwidth]{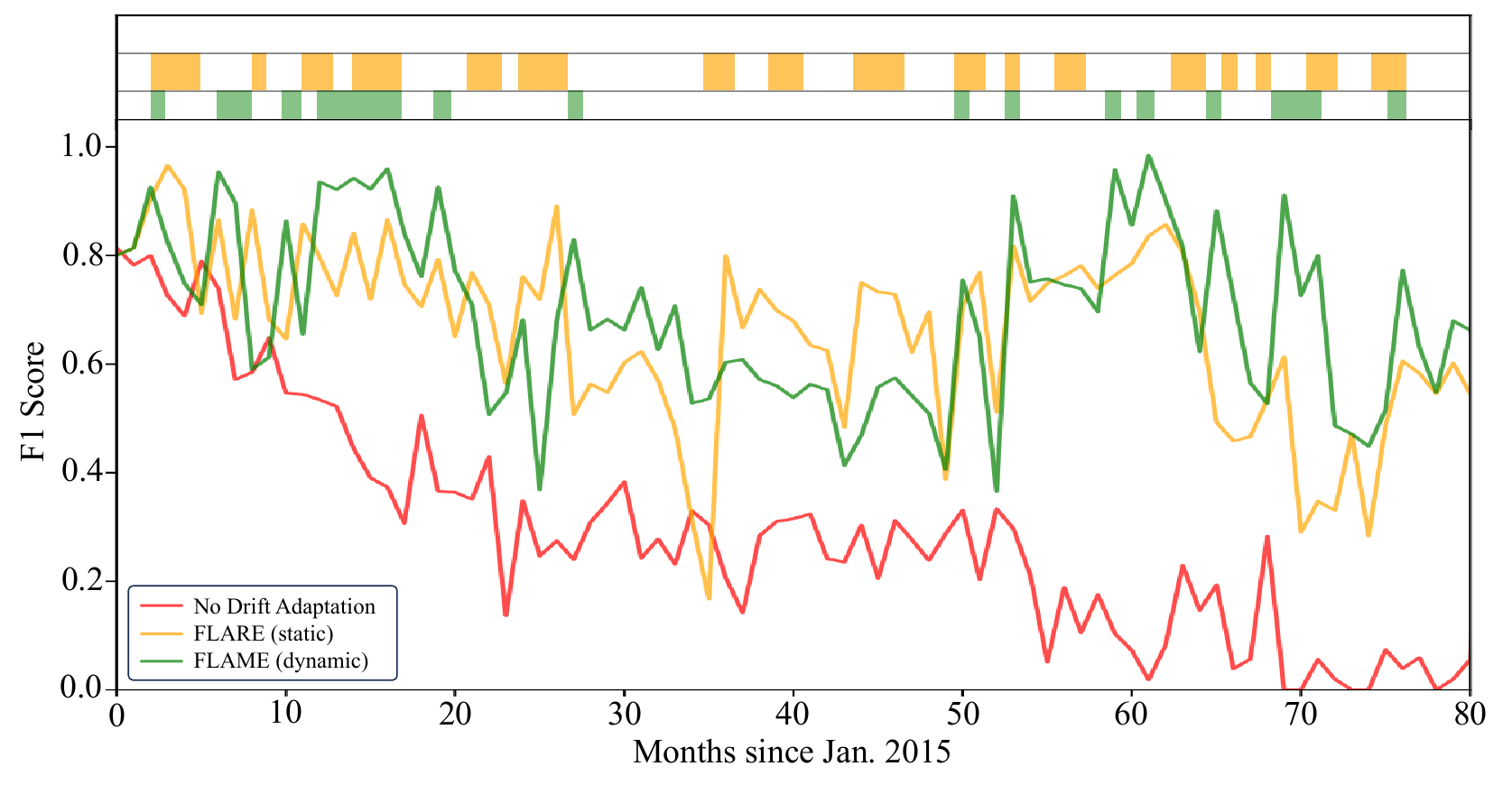}
     \vspace{-3mm}
    \caption{Small-factor experiment comparing static and dynamic thresholds.}
    \label{fig:small}
     \vspace{-3mm}
\end{figure}

We initially conducted a small-scale experiment with a single FL client and a single endpoint, using both FLARE (static thresholds) and FLAME (dynamic thresholds) methods. Around $12$k applications were selected initially for training, and $80$k applications were used for inference and retraining. Our results can be found in Fig.~\ref{fig:small}. The F1 Score significantly decreases after around ten months when drift is present. In contrast, it stays relatively high when introducing a drift detection and mitigation strategy (similar to the initial model). Comparing FLARE and FLAME, we see that they achieve roughly equal performance over time and similar trends regarding how the F1 Score increases or decreases. 

Running the experiment across many small-factor datasets, we observe that the F1 Score significantly deviates over time. This is due to the different subsets of the applications for every experiment (thus, the big dip in the performance between months 35-50 in Fig.~\ref{fig:small}). Comparing the retraining required and the data exchange, the top of Fig.~\ref{fig:small} shows the timeframes (in months) that retraining was executed. A month is considered a ``retraining month'' when retraining was triggered from the previous month's drift detection. As seen, with the dynamic thresholding and the retention of the data introduced, this time is significantly reduced, thus reducing the computational load in the system without affecting the F1 Score. This is also reflected in the total volume of data exchanged as well, where FLAME exchanged around \SI{13}{\giga\byte} with FLARE exchanging almost \SI{22}{\giga\byte}.

\subsection{Large Scale FL Experiment}

\begin{figure}[t]
    \centering
    \includegraphics[width=0.98\columnwidth]{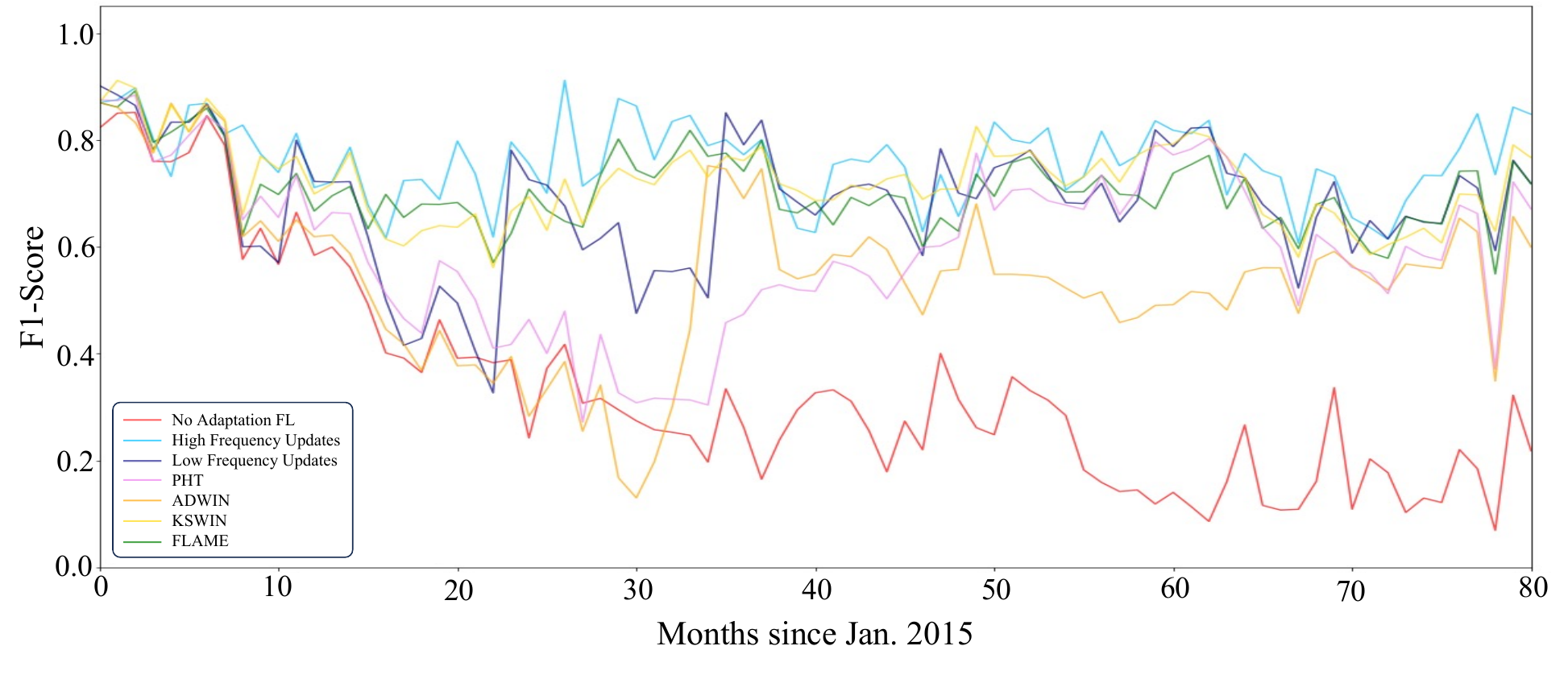}
     \vspace{-4mm}
    \caption{F1-Score of all different methods.}
    \vspace{-4mm}
    \label{fig:f1_score}
\end{figure}

\begin{figure}[t]
    \centering
    \includegraphics[width=0.98\columnwidth]{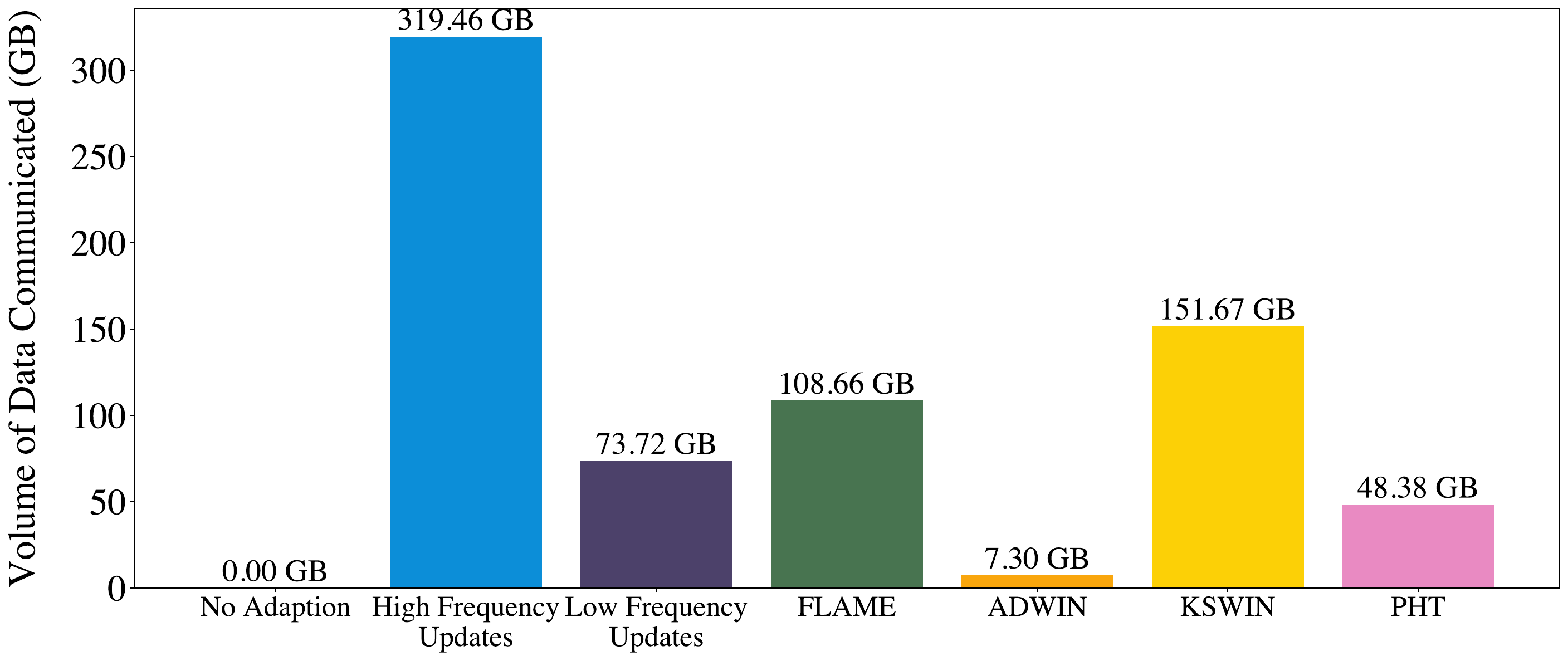}
     \vspace{-4mm}
    \caption{Volume of data exchanged across all methods.}
    \vspace{-4mm}
    \label{fig:data}
\end{figure}

As it was shown that FLAME performs better compared to FLARE, our large-scale experiment compares FLAME with a high-frequency update method (every month, the model is retrained and redeployed regardless of the performance drop), a low-frequency update method (the retraining and redeployment happen every three months) and the three concept drift detectors introduced in Sec.~\ref{subsec:detectors}. The random seed was fixed to ensure we allocate the same applications to each client and sensor across the entire experiment length and different runs. Fig.~\ref{fig:f1_score} summarises the F1 scores perceived from all methods. The no-adaptation method is our baseline, as no retraining is considered throughout the experiment.

As in the small-scale experiment, we see a significant drop of about $20\%$ in the model's performance after a few months (starting from month 6) (as seen in the baseline measurement). Moreover, we see that FLAME achieved similar performance with the high-frequency method and KSWIN while outperforming PHT, ADWIN and the low-frequency method. Each method has its own sensitivity, reflected in the number of times throughout the 80 months that drift is detected on any of the sensors. ADWIN reported few detections; thus, the retraining was very sparse. We believe this is due to the absolute values of the confidence intervals fed to the detectors (always between $0$-$1$), making ADWIN significantly underperform in such a scenario. PHT and KSWIN perform better in such a distribution, and this is reflected in the number of detections they reported and, of course, in the perceived model quality as well. 

Regarding resource utilisation, FLAME, compared to KSWIN, showed similar performance and triggered a retraining cycle around $15\%$ fewer times, achieving the same result with fewer computing resources. A similar result is also shown in the data exchanged (Fig.~\ref{fig:data}). FLAME reduced the communication overhead by $33\%$, compared to KSWIN and by almost $66\%$ compared to the high-frequency adaptation method. Overall, FLAME has managed to maintain the same performance using less computing and network resources, making it a promising solution for resource-sparse IoT environments.


\section{Conclusions}\label{sec:conclusions}
This paper presented FLAME, which effectively addresses the challenges of concept drift in large-scale IoT environments. Operating in an FL fashion, it tightly integrates within a FedOps pipeline, leveraging monitoring capabilities, intelligent data handling and various adaptations for smarter real-time operation. The framework's dynamic thresholding and data retention capabilities ensure sustained model accuracy and performance over time, even in the face of evolving data distributions. Comparative experiments highlight FLAME's ability to outperform traditional drift detection methods while minimising computational and communication resources. These findings underscore FLAME's potential as a scalable and efficient solution for maintaining the reliability of ML models in resource-constrained IoT systems. Future work will explore further optimisations and extensions of FLAME to enhance its applicability across diverse IoT applications and environments.



\bibliographystyle{ACM-Reference-Format}
\bibliography{sample-base,bib}


\begin{thebibliography}{27}


\ifx \showCODEN    \undefined \def \showCODEN     #1{\unskip}     \fi
\ifx \showDOI      \undefined \def \showDOI       #1{#1}\fi
\ifx \showISBNx    \undefined \def \showISBNx     #1{\unskip}     \fi
\ifx \showISBNxiii \undefined \def \showISBNxiii  #1{\unskip}     \fi
\ifx \showISSN     \undefined \def \showISSN      #1{\unskip}     \fi
\ifx \showLCCN     \undefined \def \showLCCN      #1{\unskip}     \fi
\ifx \shownote     \undefined \def \shownote      #1{#1}          \fi
\ifx \showarticletitle \undefined \def \showarticletitle #1{#1}   \fi
\ifx \showURL      \undefined \def \showURL       {\relax}        \fi
\providecommand\bibfield[2]{#2}
\providecommand\bibinfo[2]{#2}
\providecommand\natexlab[1]{#1}
\providecommand\showeprint[2][]{arXiv:#2}

\bibitem[Abdel~Wahab(2022)]%
        {iotDriftDetection}
\bibfield{author}{\bibinfo{person}{Omar Abdel~Wahab}.}
  \bibinfo{year}{2022}\natexlab{}.
\newblock \showarticletitle{{Intrusion Detection in the IoT Under Data and
  Concept Drifts: Online Deep Learning Approach}}.
\newblock \bibinfo{journal}{\emph{IEEE Int. Things J.}} \bibinfo{volume}{9},
  \bibinfo{number}{20} (\bibinfo{year}{2022}), \bibinfo{pages}{19706--19716}.
\newblock


\bibitem[Allix et~al\mbox{.}(2016)]%
        {androzoo}
\bibfield{author}{\bibinfo{person}{Kevin Allix} {et~al\mbox{.}}}
  \bibinfo{year}{2016}\natexlab{}.
\newblock \showarticletitle{{AndroZoo: Collecting Millions of Android Apps for
  the Research Community}}. In \bibinfo{booktitle}{\emph{Proc. of IEEE/ACM
  MSR}}. \bibinfo{pages}{468--471}.
\newblock


\bibitem[Ang et~al\mbox{.}(2013)]%
        {HAng2013}
\bibfield{author}{\bibinfo{person}{Hock~Hee Ang} {et~al\mbox{.}}}
  \bibinfo{year}{2013}\natexlab{}.
\newblock \showarticletitle{{Predictive Handling of Asynchronous Concept Drifts
  in Distributed Environments}}.
\newblock \bibinfo{journal}{\emph{IEEE Trans. Knowl. Data Eng.}}
  \bibinfo{volume}{25}, \bibinfo{number}{10} (\bibinfo{year}{2013}),
  \bibinfo{pages}{2343--2355}.
\newblock


\bibitem[Arp et~al\mbox{.}(2014)]%
        {DArp2014}
\bibfield{author}{\bibinfo{person}{Daniel Arp} {et~al\mbox{.}}}
  \bibinfo{year}{2014}\natexlab{}.
\newblock \showarticletitle{{Drebin: Effective and Explainable Detection of
  Android Malware in your Pocket.}}. In \bibinfo{booktitle}{\emph{Proc. of
  Ndss}}, Vol.~\bibinfo{volume}{14}. \bibinfo{pages}{23--26}.
\newblock


\bibitem[Bian et~al\mbox{.}(2022)]%
        {iotMLSurvey}
\bibfield{author}{\bibinfo{person}{Jiang Bian} {et~al\mbox{.}}}
  \bibinfo{year}{2022}\natexlab{}.
\newblock \showarticletitle{{Machine Learning in Real-Time Internet of Things
  (IoT) Systems: A Survey}}.
\newblock \bibinfo{journal}{\emph{IEEE Internet Things J.}}
  \bibinfo{volume}{9}, \bibinfo{number}{11} (\bibinfo{year}{2022}),
  \bibinfo{pages}{8364--8386}.
\newblock


\bibitem[Bifet and Gavaldà(2007)]%
        {adwin}
\bibfield{author}{\bibinfo{person}{Albert Bifet} {and} \bibinfo{person}{Ricard
  Gavaldà}.} \bibinfo{year}{2007}\natexlab{}.
\newblock \showarticletitle{{Learning from Time-Changing Data with Adaptive
  Windowing}}. In \bibinfo{booktitle}{\emph{Proc. of Int. Conf. SDM 2007}}.
\newblock


\bibitem[Canonaco et~al\mbox{.}(2021)]%
        {GCanonaco2021}
\bibfield{author}{\bibinfo{person}{Giuseppe Canonaco}, \bibinfo{person}{Alex
  Bergamasco}, \bibinfo{person}{Alessio Mongelluzzo}, {and}
  \bibinfo{person}{Manuel Roveri}.} \bibinfo{year}{2021}\natexlab{}.
\newblock \showarticletitle{{Adaptive Federated Learning in Presence of Concept
  Drift}}. In \bibinfo{booktitle}{\emph{Proc. of IEEE IJCNN}}.
\newblock


\bibitem[Capra et~al\mbox{.}(2019)]%
        {capra2019edge}
\bibfield{author}{\bibinfo{person}{Maurizio Capra} {et~al\mbox{.}}}
  \bibinfo{year}{2019}\natexlab{}.
\newblock \showarticletitle{{Edge Computing: A Survey on the Hardware
  Requirements in the Internet of Things World}}.
\newblock \bibinfo{journal}{\emph{Future Internet}} \bibinfo{volume}{11},
  \bibinfo{number}{4} (\bibinfo{year}{2019}), \bibinfo{pages}{100}.
\newblock


\bibitem[Casado et~al\mbox{.}(2021)]%
        {FCasado2021}
\bibfield{author}{\bibinfo{person}{Fernando~E Casado} {et~al\mbox{.}}}
  \bibinfo{year}{2021}\natexlab{}.
\newblock \showarticletitle{{Concept Drift Detection and Adaptation for
  Federated and Continual Learning}}.
\newblock \bibinfo{journal}{\emph{Multimedia Tools and Applications}}
  \bibinfo{volume}{81}, \bibinfo{number}{3} (\bibinfo{year}{2021}),
  \bibinfo{pages}{3397--3419}.
\newblock


\bibitem[{Chow} et~al\mbox{.}(2023)]%
        {flare}
\bibfield{author}{\bibinfo{person}{T. {Chow}} {et~al\mbox{.}}}
  \bibinfo{year}{2023}\natexlab{}.
\newblock \showarticletitle{{FLARE: Detection and Mitigation of Concept Drift
  for Federated Learning based IoT Deployments}}. In
  \bibinfo{booktitle}{\emph{Proc. of IWCMC}}.
\newblock


\bibitem[Herzog et~al\mbox{.}(2024)]%
        {10380759}
\bibfield{author}{\bibinfo{person}{Alexander Herzog} {et~al\mbox{.}}}
  \bibinfo{year}{2024}\natexlab{}.
\newblock \showarticletitle{{Selective Updates and Adaptive Masking for
  Communication-Efficient Federated Learning}}.
\newblock \bibinfo{journal}{\emph{IEEE Trans. Green Commun. Netw.}}
  (\bibinfo{year}{2024}), \bibinfo{pages}{1--1}.
\newblock


\bibitem[Hinkley(1971)]%
        {pageHinkley}
\bibfield{author}{\bibinfo{person}{D.~V. Hinkley}.}
  \bibinfo{year}{1971}\natexlab{}.
\newblock \showarticletitle{{Inference about the Change-point from Cumulative
  Sum Tests}}.
\newblock \bibinfo{journal}{\emph{Biometrika}} \bibinfo{volume}{58},
  \bibinfo{number}{3} (\bibinfo{year}{1971}), \bibinfo{pages}{509--523}.
\newblock


\bibitem[Jr.(1951)]%
        {ksTest}
\bibfield{author}{\bibinfo{person}{Frank J.~Massey Jr.}}
  \bibinfo{year}{1951}\natexlab{}.
\newblock \showarticletitle{{The Kolmogorov-Smirnov Test for Goodness of Fit}}.
\newblock \bibinfo{journal}{\emph{J. Amer. Statist. Assoc.}}
  \bibinfo{volume}{46}, \bibinfo{number}{253} (\bibinfo{year}{1951}),
  \bibinfo{pages}{68--78}.
\newblock


\bibitem[Kodali et~al\mbox{.}(2015)]%
        {kodali2015implementation}
\bibfield{author}{\bibinfo{person}{Ravi~Kishore Kodali},
  \bibinfo{person}{Govinda Swamy}, {and} \bibinfo{person}{Boppana Lakshmi}.}
  \bibinfo{year}{2015}\natexlab{}.
\newblock \showarticletitle{{An Implementation of IoT for Healthcare}}. In
  \bibinfo{booktitle}{\emph{Proc. of IEEE RAICS}}. \bibinfo{pages}{411--416}.
\newblock


\bibitem[{Mavromatis} et~al\mbox{.}(2023)]%
        {le3dDataDrift}
\bibfield{author}{\bibinfo{person}{I. {Mavromatis}} {et~al\mbox{.}}}
  \bibinfo{year}{2023}\natexlab{}.
\newblock \showarticletitle{{LE3D: A Lightweight Ensemble Framework of Data
  Drift Detectors for Resource-Constrained Devices}}. In
  \bibinfo{booktitle}{\emph{Proc. of IEEE CCNC}}.
\newblock


\bibitem[McLaughlin et~al\mbox{.}(2017)]%
        {NMcLaughlin2017}
\bibfield{author}{\bibinfo{person}{Niall McLaughlin} {et~al\mbox{.}}}
  \bibinfo{year}{2017}\natexlab{}.
\newblock \showarticletitle{{Deep Android Malware Detection}}. In
  \bibinfo{booktitle}{\emph{Proc. of ACM CODASPY}}. \bibinfo{pages}{301--308}.
\newblock


\bibitem[Mehmood et~al\mbox{.}(2021)]%
        {smartcities4010021}
\bibfield{author}{\bibinfo{person}{Hassan Mehmood} {et~al\mbox{.}}}
  \bibinfo{year}{2021}\natexlab{}.
\newblock \showarticletitle{{Concept Drift Adaptation Techniques in Distributed
  Environment for Real-World Data Streams}}.
\newblock \bibinfo{journal}{\emph{Smart Cities}} \bibinfo{volume}{4},
  \bibinfo{number}{1} (\bibinfo{year}{2021}), \bibinfo{pages}{349--371}.
\newblock


\bibitem[Moon et~al\mbox{.}(2024)]%
        {moon2024fedops}
\bibfield{author}{\bibinfo{person}{JiHwan Moon}, \bibinfo{person}{SeMo Yang},
  {and} \bibinfo{person}{KangYoon Lee}.} \bibinfo{year}{2024}\natexlab{}.
\newblock \showarticletitle{{FedOps: A Platform of Federated Learning
  Operations with Heterogeneity Management}}.
\newblock \bibinfo{journal}{\emph{IEEE Access}} (\bibinfo{year}{2024}).
\newblock


\bibitem[Mu and Gilmer(2019)]%
        {mnistC}
\bibfield{author}{\bibinfo{person}{Norman Mu} {and} \bibinfo{person}{Justin
  Gilmer}.} \bibinfo{year}{2019}\natexlab{}.
\newblock \showarticletitle{{MNIST-C: A Robustness Benchmark for Computer
  Vision}}.
\newblock \bibinfo{journal}{\emph{ArXiv}}  \bibinfo{volume}{abs/1906.02337}
  (\bibinfo{year}{2019}).
\newblock


\bibitem[Pendlebury et~al\mbox{.}(2019)]%
        {FPendlebury2019}
\bibfield{author}{\bibinfo{person}{Feargus Pendlebury} {et~al\mbox{.}}}
  \bibinfo{year}{2019}\natexlab{}.
\newblock \showarticletitle{{TESSERACT: Eliminating Experimental Bias in
  Malware Classification across Space and Time}}. In
  \bibinfo{booktitle}{\emph{Proc. of USENIX Sec. Symp.}}
  \bibinfo{pages}{729--746}.
\newblock


\bibitem[Qiu et~al\mbox{.}(2020)]%
        {surveyAndroozoo}
\bibfield{author}{\bibinfo{person}{Junyang Qiu} {et~al\mbox{.}}}
  \bibinfo{year}{2020}\natexlab{}.
\newblock \showarticletitle{{A Survey of Android Malware Detection with Deep
  Neural Models}}.
\newblock \bibinfo{journal}{\emph{ACM Comput. Surv.}} \bibinfo{volume}{53},
  \bibinfo{number}{6}, Article \bibinfo{articleno}{126} (\bibinfo{year}{2020}),
  \bibinfo{numpages}{36}~pages.
\newblock


\bibitem[Raab et~al\mbox{.}(2020)]%
        {kswin}
\bibfield{author}{\bibinfo{person}{Christoph Raab}, \bibinfo{person}{Moritz
  Heusinger}, {and} \bibinfo{person}{Frank-Michael Schleif}.}
  \bibinfo{year}{2020}\natexlab{}.
\newblock \showarticletitle{{Reactive Soft Prototype Computing for Concept
  Drift Streams}}.
\newblock \bibinfo{journal}{\emph{Neurocomputing}}  \bibinfo{volume}{416}
  (\bibinfo{year}{2020}).
\newblock


\bibitem[Tejesh and Neeraja(2018)]%
        {tejesh2018warehouse}
\bibfield{author}{\bibinfo{person}{B~Sai~Subrahmanya Tejesh} {and}
  \bibinfo{person}{SJAEJ Neeraja}.} \bibinfo{year}{2018}\natexlab{}.
\newblock \showarticletitle{{Warehouse Inventory Management System Using IoT
  and Open Source Framework}}.
\newblock \bibinfo{journal}{\emph{Alex. Eng. J}} \bibinfo{volume}{57},
  \bibinfo{number}{4} (\bibinfo{year}{2018}), \bibinfo{pages}{3817--3823}.
\newblock


\bibitem[Xu et~al\mbox{.}(2021)]%
        {privacy_in_ml_challenges}
\bibfield{author}{\bibinfo{person}{Runhua Xu}, \bibinfo{person}{Nathalie
  Baracaldo}, {and} \bibinfo{person}{James Joshi}.}
  \bibinfo{year}{2021}\natexlab{}.
\newblock \showarticletitle{{Privacy-Preserving Machine Learning: Methods,
  Challenges and Directions}}.
\newblock \bibinfo{journal}{\emph{CoRR}}  \bibinfo{volume}{abs/2108.04417}
  (\bibinfo{year}{2021}).
\newblock


\bibitem[Yamada and Matsutani(2023)]%
        {lightWeightConceptDrift}
\bibfield{author}{\bibinfo{person}{Takeya Yamada} {and} \bibinfo{person}{Hiroki
  Matsutani}.} \bibinfo{year}{2023}\natexlab{}.
\newblock \showarticletitle{{A Lightweight Concept Drift Detection Method for
  On-Device Learning on Resource-Limited Edge Devices}}. In
  \bibinfo{booktitle}{\emph{Proc. of IEEE IPDPSW}}. \bibinfo{pages}{761--768}.
\newblock


\bibitem[Zakeya et~al\mbox{.}(2021)]%
        {NZakeya2021}
\bibfield{author}{\bibinfo{person}{Namrud Zakeya} {et~al\mbox{.}}}
  \bibinfo{year}{2021}\natexlab{}.
\newblock \showarticletitle{{Probing AndroVul dataset for studies on Android
  malware classification}}.
\newblock \bibinfo{journal}{\emph{J. King Saud Univ., Comp. \& Info. Sci.}}
  (\bibinfo{year}{2021}).
\newblock


\bibitem[Zhong and Ge(2018)]%
        {zhong2018internet}
\bibfield{author}{\bibinfo{person}{Ray~Y Zhong} {and} \bibinfo{person}{Wenbo
  Ge}.} \bibinfo{year}{2018}\natexlab{}.
\newblock \showarticletitle{{Internet of Things Enabled Manufacturing: A
  Review}}.
\newblock \bibinfo{journal}{\emph{IJASM}} \bibinfo{volume}{11},
  \bibinfo{number}{2} (\bibinfo{year}{2018}), \bibinfo{pages}{126--154}.
\newblock


\end{thebibliography}

\end{document}